\documentclass[12pt]{article}

\usepackage{scicite}
\usepackage{amsmath}
\usepackage{amssymb}
\usepackage{times}
\usepackage{tabularx}
\usepackage{physics}
\usepackage{graphicx}
\usepackage{color}
\usepackage{soul}
\usepackage{url}

\newenvironment{sciabstract}{%
\begin{quote} \bf}
{\end{quote}}

\title{Scientific Computing with Diffractive Optical Neural Networks}

\author
{Ruiyang Chen,$^{1\dagger}$ Yingheng Tang,$^{1,2\dagger}$ Jianzhu Ma,$^{3}$  Weilu Gao,$^{1\ast}$\\
\\
\normalsize{$^{1}$Department of Electrical and Computer Engineering, University of Utah,}\\
\normalsize{Salt Lake City, UT 84112, USA}\\
\normalsize{$^{2}$Department of Electrical and Computer Engineering,  Purdue University,}\\
\normalsize{West Lafayette, IN 47907, USA}\\
\normalsize{$^{3}$Institute for AI Industry Research, Tsinghua University}\\
\normalsize{Beijing, 10084, China}\\
\normalsize{$^\dagger$These authors contribute equally.}\\
\normalsize{$^\ast$To whom correspondence should be addressed; E-mail: weilu.gao@utah.edu}
}

\date{}

\begin{document} 


\baselineskip24pt


\maketitle 


\begin{sciabstract}
  Diffractive optical neural networks (DONNs) have been emerging as a high-throughput and energy-efficient hardware platform to perform all-optical machine learning (ML) in machine vision systems. However, the current demonstrated applications of DONNs are largely straightforward image classification tasks, which undermines the prospect of developing and utilizing such hardware for other ML applications. Here, we numerically and experimentally demonstrate the deployment of an all-optical reconfigurable DONNs system for scientific computing, including guiding two-dimensional quantum material synthesis, predicting the properties of nanomaterials and small molecular cancer drugs, predicting the device response of nanopatterned integrated photonic power splitters, and the dynamic stabilization of an inverted pendulum with reinforcement learning. Despite a large variety of input data structures, we develop a universal feature engineering approach to convert categorical input features to the images that can be processed in the DONNs system. Our results open up new opportunities of employing DONNs systems for a broad range of ML applications.

\end{sciabstract}

\section*{Introduction}

Machine learning (ML) has demonstrated the state-of-the-art performance in a variety of applications, such as computer vision~\cite{LeCunEtAl2015N,GoodfellowEtAl2016}, medicine~\cite{TopolEtAl2019NM}, finance~\cite{DixonEtAl2020}, autonomous engineering design~\cite{MirhoseiniEtAl2021N}, and scientific computing~\cite{ButlerEtAl2018N,SeniorEtAl2020N}, but performing ML tasks on hardware systems requires substantial energy and computational resources. The fundamental quantum mechanics limit leads to a bottleneck of reducing the energy consumption and simultaneously increasing the integration density of electronic circuits to catch up with the increasing scale of modern large-scale ML models~\cite{TheisEtAl2017CSE,LeisersonEtAl2020S}. Optical architectures are emerging as promising high-throughput and energy-efficient ML hardware accelerators by leveraging the parallelism and low static energy consumption of a fundamentally different particle, photon, for computing~\cite{WetzsteinEtAl2020N,MenguEtAl2022AOP}. 

Among optical systems, free-space diffractive optical neural networks (DONNs) that are able to host millions of computing neurons and form deep neural network architectures can optically perform ML tasks through the spatial light modulation and optical diffraction of coherent light in multiple diffractive layers~\cite{LinEtAl2018S,LiEtAl2019AP,LuoEtAl2019LSA,MenguEtAl2019IJSTQE,MenguEtAl2020P,LiEtAl2021SR,LeonardEtAl2021P,FuEtAl2021OE,RahmanEtAl2021LSA,ChenEtAl2021E,GoiEtAl2021LSA,ZhouEtAl2021NP,LeonardEtAl2022OE,ZengEtAl2022OE,LiEtAl2022OE,LuoEtAl2022LSA,WuEtAl2022OE,LouEtAl2023OL,ChenEtAl2022LPR}.  Prior demonstrations have shown the capability of DONNs systems to recognize input images directly in optical domain. Hence, employing DONNs instead of conventional imaging optical components in machine vision systems can reduce backend electronic processing burden. However, the demonstrated applications of DONNs systems are largely straightforward image classification tasks, which undermines the prospect of developing and utilizing such hardware for other ML applications. 

Here, we numerically and experimentally demonstrate the deployment of an all-optical reconfigurable DONNs system for high-throughput and energy-efficient scientific computing, including guiding two-dimensional (2D) quantum material synthesis, predicting the properties of 2D quantum materials and cancer drugs, predicting the device response of nanopatterned integrated photonic power splitters, and the dynamic stabilization of an inverted pendulum with reinforcement learning (RL). Despite a large variety of input data structures in these diverse applications, we develop a universal feature engineering approach to convert categorical input features to binary images that can be processed in the DONNs system. In contrast to other DONNs systems, the trained models can be directly deployed to electrically controlled diffractive layers in our DONNs system so that the system can be on-demand reconfigured for different ML tasks instantaneously. Our results open up new opportunities of employing DONNs systems for a broad range of ML applications.

\section*{Results}

Figure\,1 and Fig.\,S1 show the schematic and photo of the all-optical reconfigurable DONNs experimental setup that is recently demonstrated by us in Ref.\,\cite{ChenEtAl2022LPR}. The system consists of four cascaded liquid-crystal spatial light modulators (SLMs) with the first SLM encodes input images and the other three SLMs form reconfigurable diffractive layers. Multiple polarizers and half-waveplates are employed to configure the operation modes of SLMs. Specifically, the input image SLM mainly operates in intensity modulation mode and images are formed in the binary light intensity transmittance under coherent visible laser illumination. The SLMs for diffractive layers mainly operate in phase modulation mode and the grey levels of each diffractive pixel in all diffractive layers are optimized so that coherent input images can be converged to one of predefined regions on a detector array. Each detector region can represent a variety of categorical output features in different application scenarios. More details on the experimental setup can be found in \emph{Materials and Methods}. 

Our system has several unique advantages. Through the system-specific diffraction modeling, device-specific physics-aware training, and precise optical alignment, the \emph{in silico} trained models can be directly and accurately deployed to the hardware system. Furthermore, all input images and diffractive layers represented by SLMs can be \emph{in-situ} adjusted through external electrical control signals, so that the whole system can be tailored to different ML tasks instantaneously without rebuilding or modifying any physical components. 

\subsection*{2D quantum material synthesis}

Although 2D quantum materials have been under extensive research because of their unique properties~\cite{WangEtAl2012NN,NovoselovEtAl2016S,SchaibleyEtAl2016NRM,GibertiniEtAl2019NN} and chemical vapor deposition (CVD) has become a scalable and controllable synthesis approach~\cite{ShiEtAl2015CSR}, exploring the overwhelming parameter space in CVD process to find optimal synthesis conditions leads to numerous empirical, highly uncertain, time-consuming, and expensive experimental trials. Recently, ML models have shown great potential to substantially accelerate the exploration~\cite{LiuEtAl2017JM,ButlerEtAl2018N,Correa-BaenaEtAl2018J,WeiEtAl2019I} and we employ the DONNs system to predict whether monolayer molybdenum disulfide (MoS$_2$), one type of 2D quantum materials, can be synthesized given certain process parameters.

Figure\,2A displays the schematic of the CVD synthesis of monolayer MoS$_2$ with multiple synthesis process parameters, such as feed gas flow and heating temperature profile. We utilize the experimental dataset reported in Ref.\,\cite{TangEtAl2020MT}. There are seven synthesis process parameters, which include both discrete numerical and boolean values. For example, the ramp time has 13 discrete values and whether sodium chloride salt is added is presented as True or False. To represent these categorical process parameters as the input images for the DONNs system, we develop a one-hot feature engineering approach. The values of each parameter is expressed as a vector with only one element showing one and the rest showing zero, and the vectors for all parameters are concatenated as a 2D matrix. The obtained matrix is further reshaped and resized to a 100$\times$100 image. A threshold-based binarization is applied on resized images for the DONNs system processing. The pixels with the value one in obtained images correspond to the largest transmittance in the input SLM and those with the value zero correspond to the smallest transmittance. The threshold is adjusted for different applications to have enough light transmission and detector signal-to-noise ratio. The obtained images are processed by the DONNs system and two regions on the camera represent two categorical output values, which are ``can grow (Yes)'' and ``cannot grow (No)''. 

The input images produced from the training dataset through our developed feature engineering approach are used to train the SLM diffractive layers in the DONNs system. Figure\,2B shows the simulation camera output image, intensity distribution in two camera regions, and confusion matrix using the images from the test dataset. More details on the one-hot input feature image processing for training and test datasets can be found in \emph{Materials and Methods}. The simulation results confirm the feasibility of employing the DONNs system to guide MoS$_2$ synthesis and the obtained accuracy 82\,\% is comparable to the accuracies obtained using standard ML models, such as multilayer perceptron (MLP) and XGBoost decision tree~\cite{TangEtAl2020MT}. Figure\,2C shows the corresponding experimental results, which further validate and agree well with simulation results. More data of simulation and experimental results can be found in Fig.\,S2. 

\subsection*{Material and molecule properties}

The first-principle calculations to solve Schr\"{o}dinger equations, such as density-functional theory (DFT), provide a powerful tool for obtaining accurate structure-property relationship of materials. However, such calculations are time-consuming and computationally intensive, and thus searching for suitable materials for certain applications in an astronomical chemical space is extremely challenging. Smart navigation by leveraging ML models~\cite{SchlederEtAl2019JPM,HimanenEtAl2019AS} and established databases, such as for 2D quantum materials~\cite{HaastrupEtAl2018M,ChoudharyEtAl2020CM}, can accelerate the exploration process. Specifically, we employ the DONNs system to predict the properties of 2D quantum materials in the Computational 2D Materials Database (C2DB) library~\cite{HaastrupEtAl2018M}. One input branch shown in Fig.\,3A depicts the workflow. The input features are limited to atomic structure information, such as constitute atoms, and any features calculated from DFT are explicitly excluded. These input features are converted to the input images for the DONNs system through the one-hot feature engineering approach as described before. 

Figure\,3B summarizes the simulation and experimental output images and corresponding intensity distributions of utilizing the DONNs system to predict whether 2D quantum materials are stable or not, are direct bandgap materials or not, and magnetic or not, when atomic structures are given. The numbers of materials with distinct properties are balanced in both training and test datasets. These output labels are created following Ref.\,\cite{SorkunEtAl2020CM}. Three different DONNs models are trained for predicting three properties. More details about the setup of training and test input images and output labels can be found in \emph{Materials and Methods}. The obtained accuracies from simulations for three properties are 86\,\%, 73\,\%, and 78\,\%, respectively. These values are comparable to those obtained using standard MLP models as reported in Ref.\,\cite{SorkunEtAl2020CM}. The experimental results in Fig.\,3B further validate the simulation results, although with slightly lower accuracies. More simulation and experimental data, as well as confusion matrices can be found in Figs.\,S3 and S4.

Furthermore, one of the most striking findings of cancer biology is the extreme genetic heterogeneity among cancer patients. The heterogeneity of tumor genomes poses a fundamental challenge for choosing the cancer drugs in the clinic. Recently, ML has been promising in predicting the clinical response of cancer drugs only based on the patients' genome mutation~\cite{KuenziEtAl2020CC,HuangEtAl2021GB,ZhuEtAl2021NB}. We employ the DONNs system for the prediction of such biological datasets. Specifically, we focus on three experimental datasets of small molecular drugs, including PD0325901, Refametinib, and Selumetinib~\cite{IorioEtAl2016C}. They are all MEK inhibitors that target at different cancer types and have different side effects. Instead of directly converting the input features of genome mutation into the input images for the DONNs system, a feature reduction technique is first utilized as shown in Fig.\,3A. More details can be found in \emph{Materials and Methods}. 

Figure\,3C summarizes the simulation and experimental results of utilizing the DONNs system to predict whether the PD0325901 cancer inhibitor are effective or not for different tumor cell lines. The simulation and experimental results for the other two drugs, Refametinib and Selumetinib, are presented in Figs\,S5 and S6. In addition, confusion matrices for all three drugs are shown in Fig\,S7. The training and test datasets are balanced. More details on the model setup can be found in \emph{Materials and Methods}. The results clearly show the feasibility of employing the DONNs system to predict cancer drug effectiveness, and simulation and experimental results match well.

We would like to highlight that all trained DONNs models can be instantaneously loaded into the DONNs system because of the system reconfigurability without the need of modifying any physical components and adaptive tuning~\cite{ZhouEtAl2021NP}. Although the used liquid-crystal SLM for encoding input images in our DONNs system only have the refresh rate 60\,Hz and the used CMOS camera has the frame rate $<$40\,frames per second, the system throughput can be substantially improved with device innovation. For example, an electro-optic SLM based on organic molecules can achieve $>$GHz switching speed~\cite{Benea-ChelmusEtAl2022NC} and an ultrafast camera can achieve a trillion frames per second~\cite{KimEtAl2020SA}. With these devices, the DONNs system can potentially predict the properties of $>$10$^9$ materials and molecules per second for high-throughput screening.

\subsection*{Photonic device}

Similar to the time-consuming DFT calculations, the fullwave electromagnetic calculations to solve Maxwell's equations to explore structure-property relationship of photonic devices are also time-consuming and computationally intensive. The use of ML models is also beneficial to accelerate the exploration~\cite{YaoEtAl2019N,JiangEtAl2021NRM,MaEtAl2020NP,KudyshevEtAl2020P}. We demonstrate that the DONNs system can also be used to predict device response given its topology. Figure\,4A shows the schematic of a nanopatterned integrated photonic power splitter, which splits the power from the input tapered waveguide ($P_0$) into the power in two output tapered waveguides ($P_1$ and $P_2$). The splitter is manufactured on a silicon-on-insulator substrate. The center component connecting three waveguides consists of an array of 20$\times$20 grid points. At each point, there is either a hole of various diameters or no hole, which defines the device topology and the power splitting ratio $P_1/P_2$. 

We resize the device topology to match the size of input images for the DONNs system, which predicts the power splitting ratio with three categorical output labels $5:5$, $7:3$, and $9:1$. The datasets are generated from a conditional variational autoencoder trained from a dataset calculated using fullwave electromagnetic simulations as described in Ref.\,\cite{TangEtAl2020LPR}. More details about the input image mapping and dataset generation can be found in \emph{Materials and Methods}. Figures\,4B and 4C show the camera output images, intensity distributions, and confusion matrices from simulations and experiments, respectively. The obtained prediction accuracies are close to 100\,\%. More simulation and experimental data can be found in Fig\,S8. 

\subsection*{Inverted pendulum}

An inverted pendulum is the pendulum with its center of mass above the pivot point, which is unstable. The pivot point can be horizontally moved back and forth through a control feedback loop to keep it balanced at the inverted position and keep it from falling. The motion evolution and dynamic stabilization of an inverted pendulum is a canonical problem of dynamics and control theory. Not only in classical mechanics, quantum mechanical systems, such as a many-body spin system described in Ref.\,\cite{HoangEtAl2013PRL}, have similar dynamic stabilization of a quantum inverted pendulum. Recently, RL approaches, such as deep Q-learning, provide fresh perspectives of solving physics tasks~\cite{LillicrapEtAl2015APA} and we demonstrate that the DONNs system can be used in a RL framework to dynamically stabilize an inverted pendulum.

Figure\,5A shows the schematic of the RL framework. Four states, including the positions and accelerations of the pendulum and base, are encoded into the input images of the DONNs using the one-hot feature engineering approach described before. The action space consists of moving the base left or right. The DONNs system is trained to take the current state as input and generate the next-step action as output so that the pendulum can be kept stable as many steps as possible. Hence, the reward function is defined as the number of steps to keep the pendulum from falling and we choose 200 as the number of steps to stop the training process. More details on the training of the DONNs system can be found in \emph{Materials and Methods}. The trained DONNs can generate actions to keep the pendulum stable for 200 steps. Figure\,5B displays two example sets of pendulum images, DONNs input state images, simulation action camera output images, and experimental camera action output images, which confirm the feasibility of utilizing the DONNs system for RL applications. The step evolution of these four images is displayed in \emph{Supplementary Movie 1}. 

\section*{Discussion}

We have numerically and experimentally demonstrated that the DONNs system can be deployed to accelerate the execution of ML models in scientific computing domains, including guiding 2D quantum material synthesis, predicting the properties of 2D quantum materials and cancer inhibitors, predicting the device response of nanopatterned integrated photonic power splitters, and the dynamic stabilization of an inverted pendulum with a RL approach. We have developed a universal feature engineering approach to convert categorical input features to the images that can be processed in the DONNs system. One limitation of current hardware system is the binary nature of input images and future work can leverage the greyscale encoding to handle complicated categorical and even continuous input features. 


\section*{Materials and Methods}

\noindent\emph{\underline{Experimental setup.}} The input light is from a laser diode (CPS532 from Thorlabs, Inc.), which has the center wavelength at 532\,nm and a beam diameter $\sim$4\,mm. The distance between SLMs and the camera is 11\,inches. The model number of all transmissive SLMs is LC 2012 from HOLOEYE Photonics AG. The refresh rate is 60\,Hz. The output images are captured by a CMOS camera (CS165MU1 from Thorlabs, Inc.). The frame rate is 34.8\,frames per second and 10 frames are averaged to obtain output images. The input image size is 100$\times$100. More experimental details can be found in our recent work~\cite{ChenEtAl2022LPR}. 

\noindent\emph{\underline{2D quantum material synthesis.}} The dataset contains a total of 300 MoS$_2$ synthesis data. There are seven input features in the dataset, which are the distance of sulfur outside furnace, flow rate, reaction temperature, ramp time, reaction time, whether the substrate is placed flat or tilted, and whether sodium chloride salt is added or not. The first five features contain discrete real-valued numbers and the last two features are Boolean. The detailed data structure can be found in the open-source repository~\cite{github_repo1}. The output label is binary with 0 representing unsuccessful growth (small flake size) and 1 representing successful growth (large flake size). We randomly select 240 data (80\,\% data) as the training set and the rest (20\,\% data) as the test set. The input image size of our DONNs system is 100$\times$100 with binary grey levels and we convert input features to DONNs input images in following steps. First, we encode the input features following one-hot fashion, where the number of input features is changed from 7 to 81. The 1$\times$81 vector is reshaped to a 9$\times$9 matrix. This binary matrix is further scaled up through interpolation to a greyscale 100$\times$100 matrix, which finally becomes binary again by setting a threshold. In this application, the threshold is set as 0.1.

\noindent\emph{\underline{2D quantum material properties.}} For the prediction of all three properties, the input features are the same, which are the number of atoms per unit cell, prototype vector, and chemical composition vector. The output labels to represent ``magnetic state'', ``is stable'', and ``has direct band gap'' are available in the dataset. The feature extraction from materials is the same with Ref.\,\cite{SorkunEtAl2020CM} and the details can be found in the open-source repository~\cite{github_repo2}. The similar one-hot feature engineering process as described in the previous section is applied to 1$\times$144 vectors, which are further converted to input images of the size 100$\times$100 for the DONNs system through reshaping, resizing, and threshold-binarization processing. The threshold is set as 0.45. We split the data into the training and test data with a ratio of approximately 4:1. For the prediction of magnetic state, the training set has 952 data and the test set has 202 data. For the prediction of stability, the training set has 800 data and the test set has 208 data. For the prediction of band structure, the training set has 220 data and the test set has 52 data.

\noindent\emph{\underline{Cancer inhibitor effectiveness.}} For the effectiveness prediction of cancer inhibitors, including PD0325901, Refametinib, and Selumetinib, each drug has totally 7798 binary input features of patients' genomes mutation. Because of the limitation in the DONNs hardware system, we reduce the number of features by using a linear regression model with L1 normalization to extract the first 144 features. We then convert these features to a 100$\times$100 input image for the DONNs system by reshaping a 1$\times$144 vector to a 12$\times$12 matrix, resizing the 12$\times$12 matrix to a 100$\times$100 matrix through interpolation, and threshold-binarization processes. The threshold is set as 0.5. The output labels are denoted as real-valued numbers in a range from 0 to 1, which represent effectiveness of cancer inhibition. They become binary with a threshold 0.6, meaning that any value greater than 0.6 represents ``effective'' and other values represent ``ineffective''. We split the data into the training and test data with a ratio of approximately 4:1. For PD0325901, there are 1465 training data and 349 test data. For Refametinib, there are 1329 training data and 355 test data. For Selumetinib, there are 1482 training data and 328 test data.

\noindent\emph{\underline{Integrated photonic beam splitter.}} The center component has an area of 2.25$\times$2.25\,$\mu\mathrm{m}^2$. Hole spacings are 112\,nm from center to center of each hole position, and the maximum and minimum hole diameters are 77\,nm and 42\,nm. If a diameter is below 42\,nm, there is no hole. A conditional variational autoencoder (CVAE) is trained using the simulation data generated fullwave electromagnetic finite-difference-time-domain calculations as described in detail in Ref.\,\cite{TangEtAl2020LPR}. Here, we use the decoder of trained CVAE to generate the device topology given the splitting ratio. For each ratio, we generate 800 training data and 131 test data. The device topology is represented by a 20$\times$20 matrix and each element is a floating point number to represent a normalized hole size. The 20$\times$20 matrix is then scaled up through interpolation to a 100$\times$100 matrix, which then becomes binary with the threshold 0.5.

\noindent\emph{\underline{Inverted pendulum stabilization.}} Four states are the horizontal position of the base in a range from -4.8 to 4.8, the angle between the pendulum and the vertical position in a range from -0.418 to 0.418, the horizontal velocity of the cart in a range from $-\infty$ to $+\infty$, and the angular velocity of the pendulum in a range from $-\infty$ to $+\infty$. The values of first two features are normalized to a range from 0 to 1 through linear normalization, and the latter two are normalized to the same range through the sigmoid function. The normalized range is then discretized to 25 levels, which is also the dimension of one-hot encoded feature vectors. Similar to other demonstrations, the 100$\times$100 input feature images for the DONNs system are generated through reshaping, resizing interpolation, and binarization processes. The threshold is set as 0.01. 

In order to facilitate the training of the DONNs system, we first train a convolutional neural networks (CNNs) model in the RL framework shown in Fig.\,5A. The CNNs model has the state images generated using the one-hot feature engineering approach as input and the next-step action as the output. The CNN model contains three convolutional layers with the channel size of 1, 16, and 32, respectively. The kernel size is 5 and the step size is 2. Each convolutional layer is followed by a batch normalization layer. After the last convolutional layer, a fully connected layer is included to reduce the feature dimension to 2 to represent two actions. The reward function is defined as the number of steps to have a stabilized pendulum and we choose 200 steps as the training stop criterion. When the CNNs model is trained, it is used to perform the inference of multiple inverted pendulum problems with different initial states, and all steps with input images and output action are used to train the DONNs system.

\bibliographystyle{Science}

\newpage
\section*{Acknowledgments}

\noindent\emph{\underline{Funding:}} R.\,C. and W.\,G. thank the support from the University of Utah start-up fund.

\noindent\emph{\underline{Author contributions:}} W.\,G. conceived the idea, and designed and supervised the project. R.\,C. performed all experiments and Y.\,T. developed machine learning algorithms under the guidance of W.\,G. J.\,M. helped with the development of machine learning algorithm for cancer drug prediction. All authors wrote and discussed the manuscript.

\noindent\emph{\underline{Competing interests:}} The authors declare that they have no competing financial interests.

\noindent\emph{\underline{Data and materials availability:}} All data needed to evaluate the conclusions in the paper are presented in the paper and/or the Supplementary Materials. 
 
\section*{Supplementary materials}
Figs. S1 to S8\\
Supplementary Movie 1\\

\clearpage

\begin{figure}
  \centering
  \includegraphics[width=\textwidth]{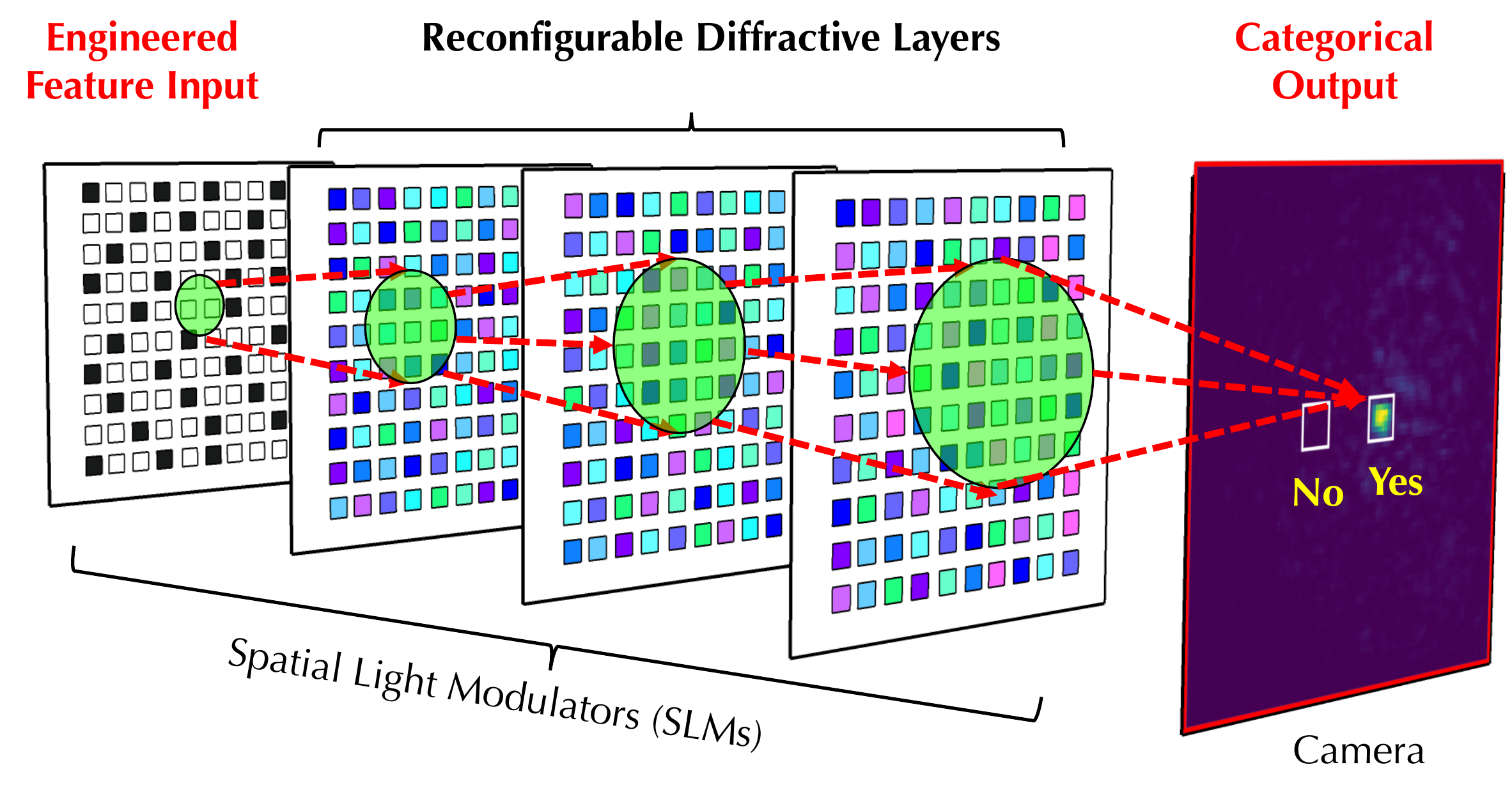}
\end{figure}

\noindent {\bf Fig.\,1. Schematic of the all-optical reconfigurable DONNs system.} The first (leftmost) SLM encodes input images that are generated from input features by one-hot feature engineering. The second to forth SLMs form three reconfigurable diffractive layers. Input images generated by shining coherent light can be converged into multiple predefined regions on the output camera due to spatial light modulation and optical diffraction from optimized diffraction layers. Each detector region represents a categorical output label. 

\newpage 

\begin{figure}
  \centering
  \includegraphics[width=\textwidth]{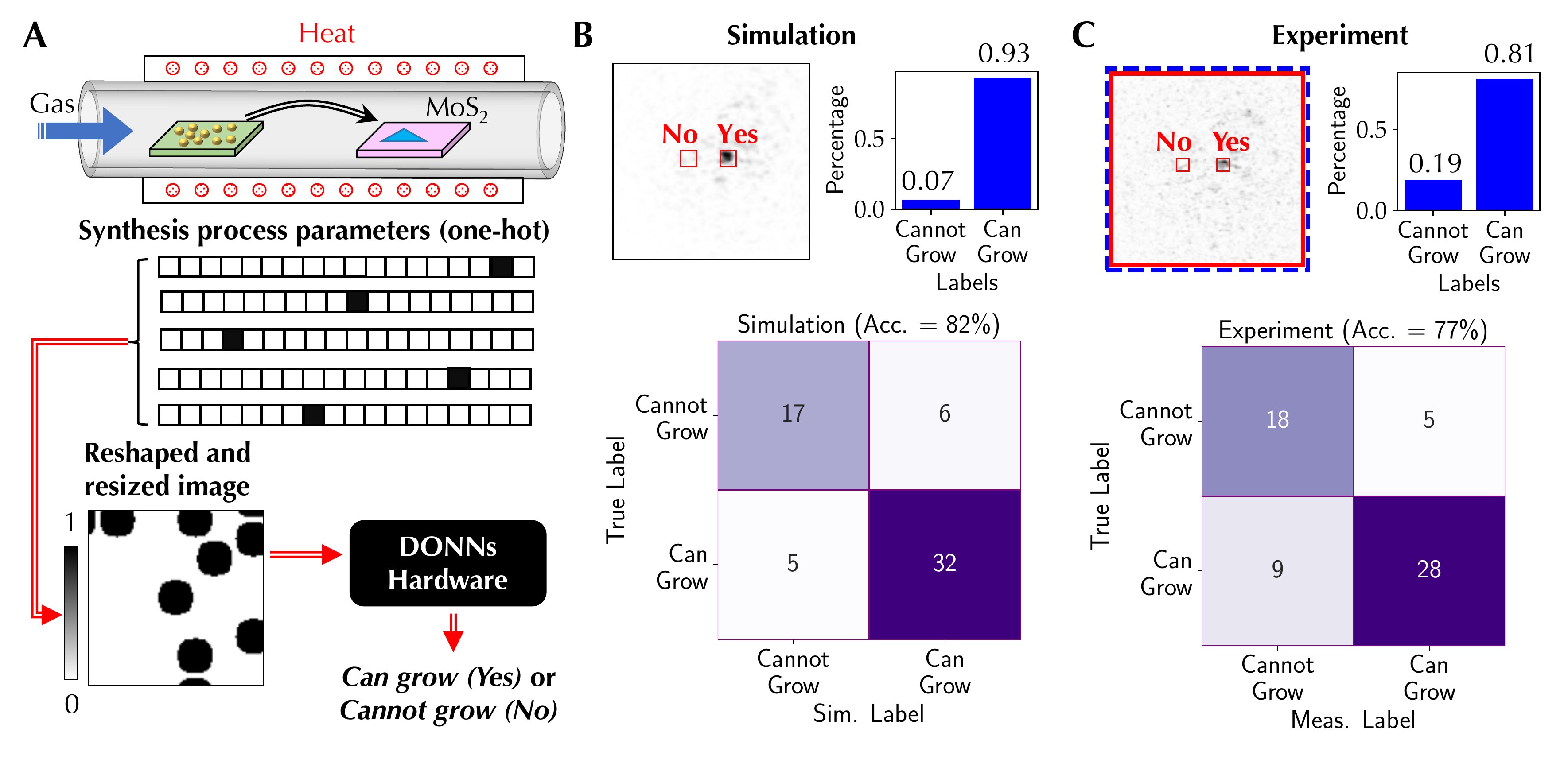}
\end{figure}

\noindent {\bf Fig.\,2. 2D quantum materials synthesis.} \textbf{(A)}\,Schematics of the CVD synthesis of MoS$_2$ monolayer flakes and the one-hot feature engineering of categorical input process parameters. The reshaped and resized images can be processed in the DONNs system to predict whether MoS$_2$ can be grown or not given process parameters. \textbf{(B)}\,Simulation and \textbf{(C)}\,experimental camera output images, intensity distributions in two camera regions, and confusion matrices.

\newpage 

\begin{figure}
  \centering
  \includegraphics[width=\textwidth]{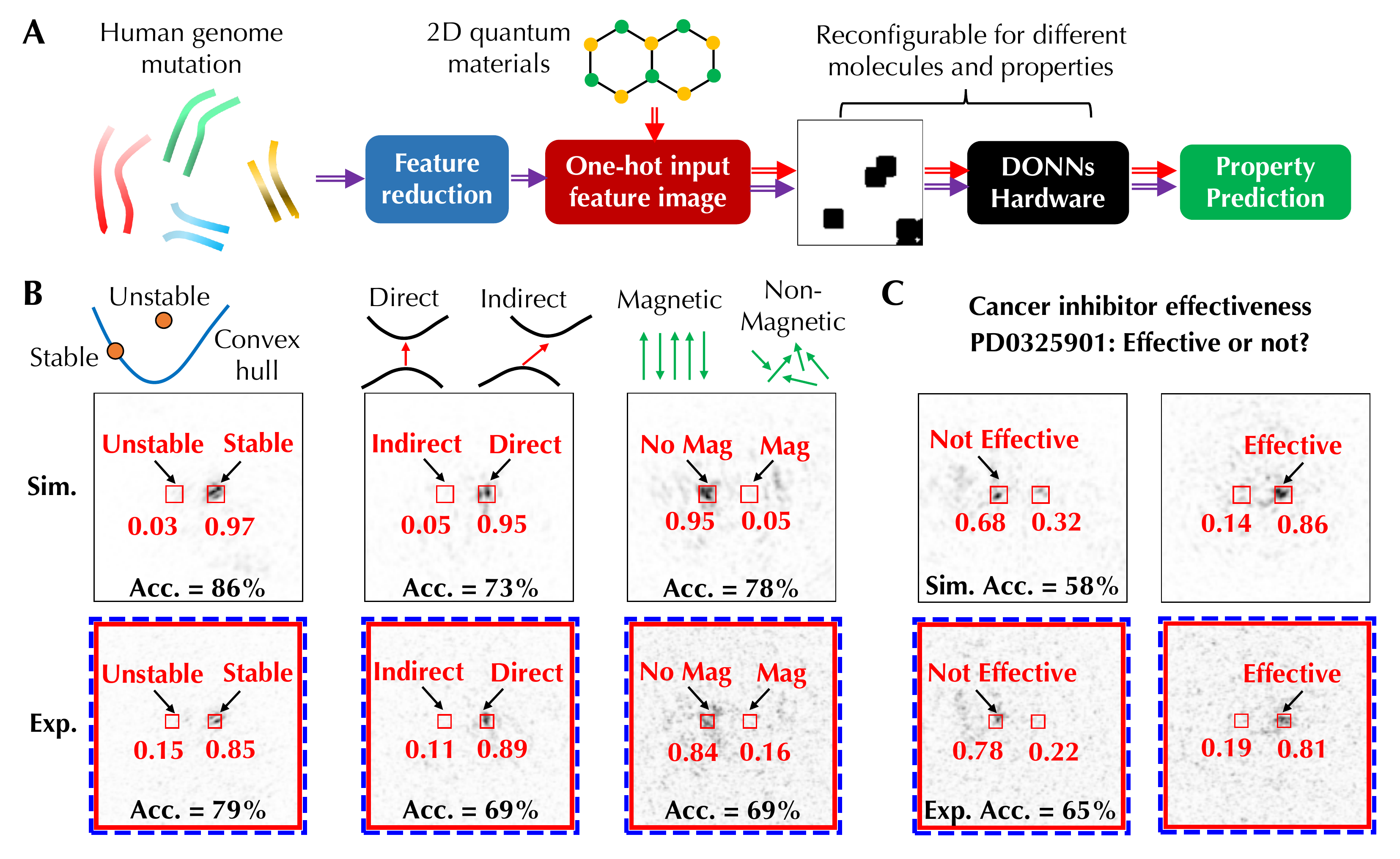}
\end{figure}

\noindent {\bf Fig.\,3. Material and molecule properties.} \textbf{(A)}\,Schematic of the workflow of predicting the properties of 2D quantum materials and cancer drugs using the DONNs system. The input features of 2D quantum materials can be directly converted to input feature images. The input features of human genome mutation information that are used to predict cancer inhibitor effectiveness are first reduced through a feature reduction technique. The diffractive layers can be \emph{in-situ} reconfigured to screen the different properties of various materials and molecules in a high-throughput manner. \textbf{(B)}\,Simulation and experimental camera output images and corresponding intensity distributions for predicting three properties, including the stability, band structures, and magnetism of 2D quantum materials in the C2DB library. \textbf{(C)}\,Simulation and experimental camera output images and corresponding intensity distributions for predicting the effectiveness of PD0325901 cancer inhibitor. 

\newpage 

\begin{figure}
  \centering
  \includegraphics[width=\textwidth]{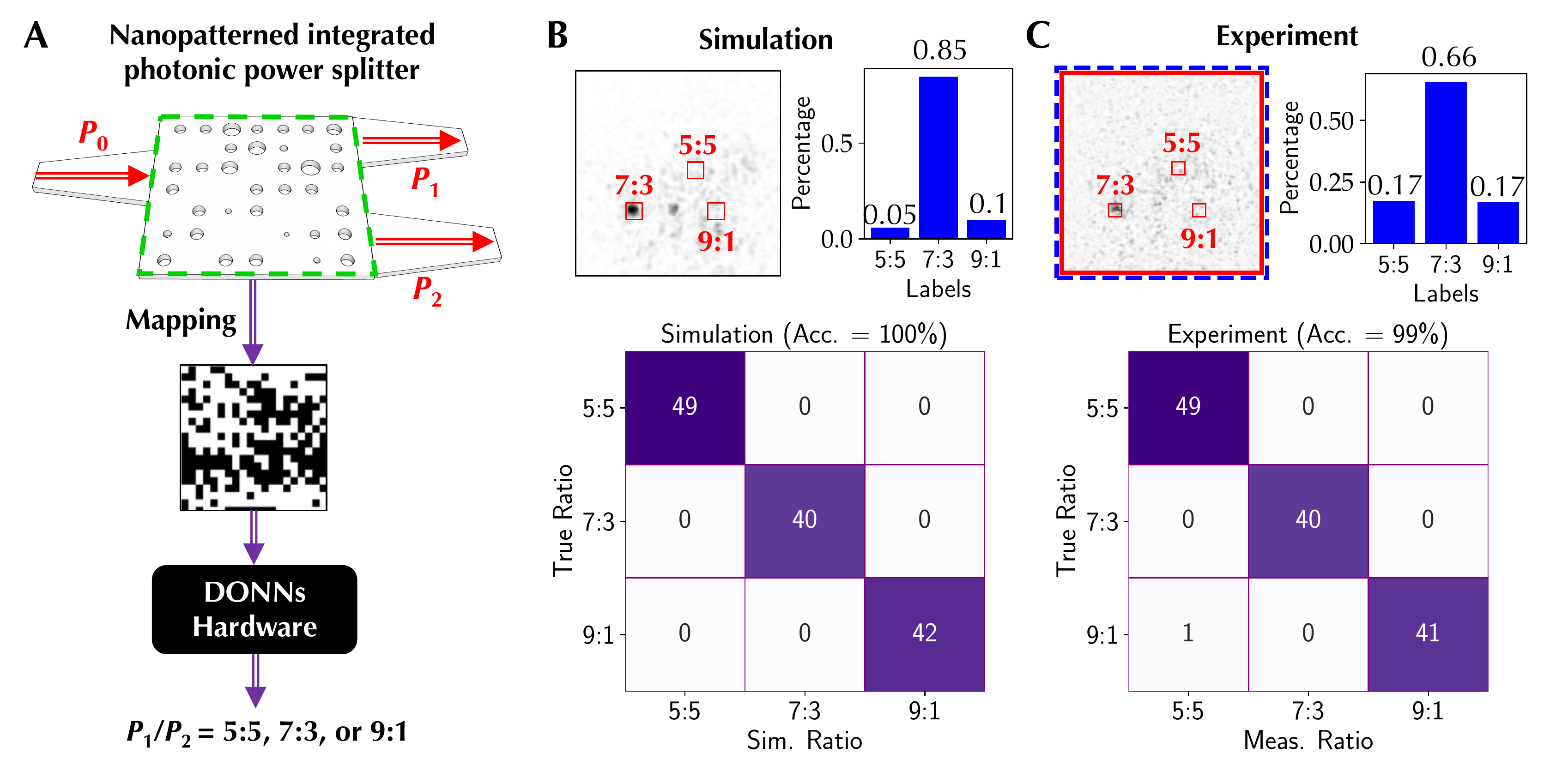}
\end{figure}

\noindent {\bf Fig.\,4. Photonic device.} \textbf{(A)}\,Schematic of the workflow of predicting the device response of nanopatterned integrated photonic power splitter on a silicon-on-insulator platform. The hole array device topology is mapped to input images. There are three output power splitting ratios, including $5:5$, $7:3$, and $9:1$. \textbf{(B)}\,Simulation and \textbf{(C)}\,experimental camera output images, intensity distributions in two camera regions, and confusion matrices. 

\newpage 

\begin{figure}
  \centering
  \includegraphics[width=\textwidth]{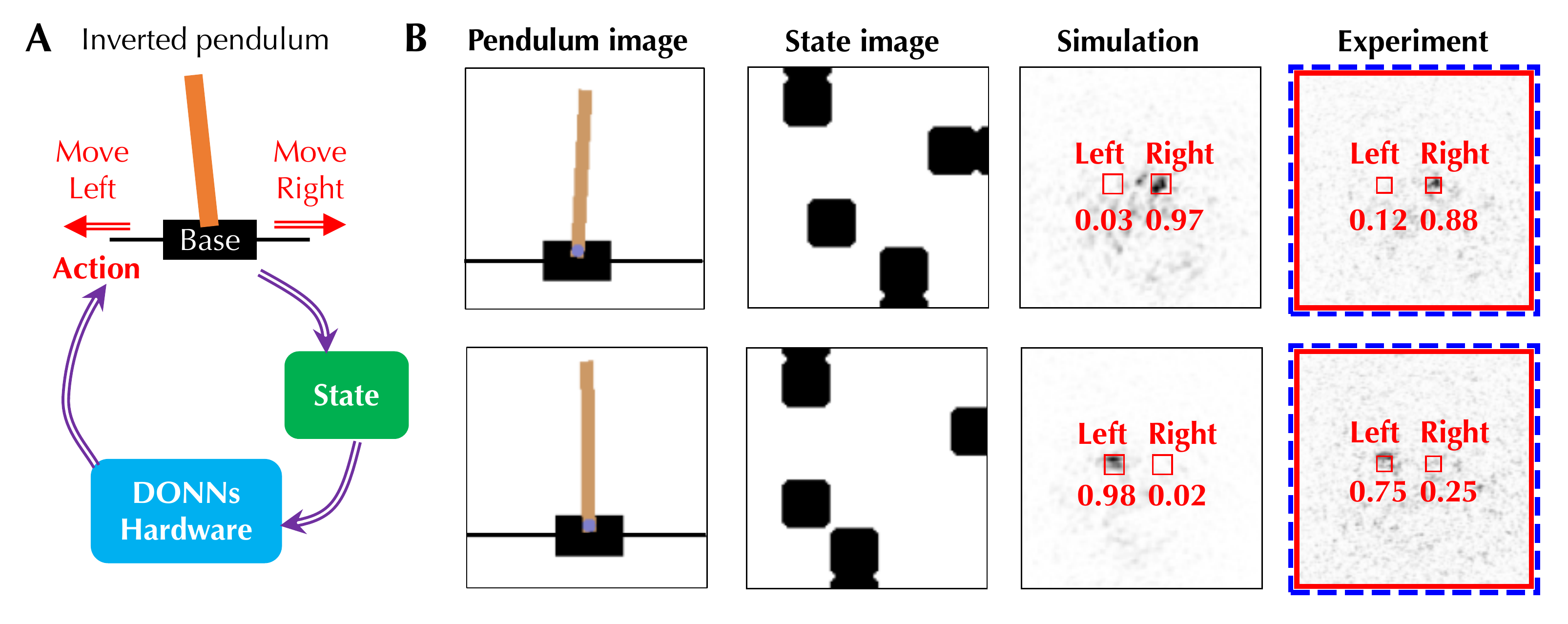}
\end{figure}

\noindent {\bf Fig.\,5. Dynamic stabilization of an inverted pendulum through RL.} \textbf{(A)}\,Schematic of the RL framework to dynamically stabilize the inverted pendulum utilizing the DONNs system. There are four states including the positions and accelerations of the pendulum and base. The action space consists of moving the base left or right. \textbf{(B)}\,Pendulum images, DONNs input feature images, simulation action camera output images, and experimental camera action output images for two different actions. 


\end{document}



\baselineskip24pt


\maketitle 

\pagebreak

\noindent\textbf{Supplementary Figures S1 - S8}
\begin{figure}[h]
	\begin{center}
		\includegraphics[width=\textwidth]{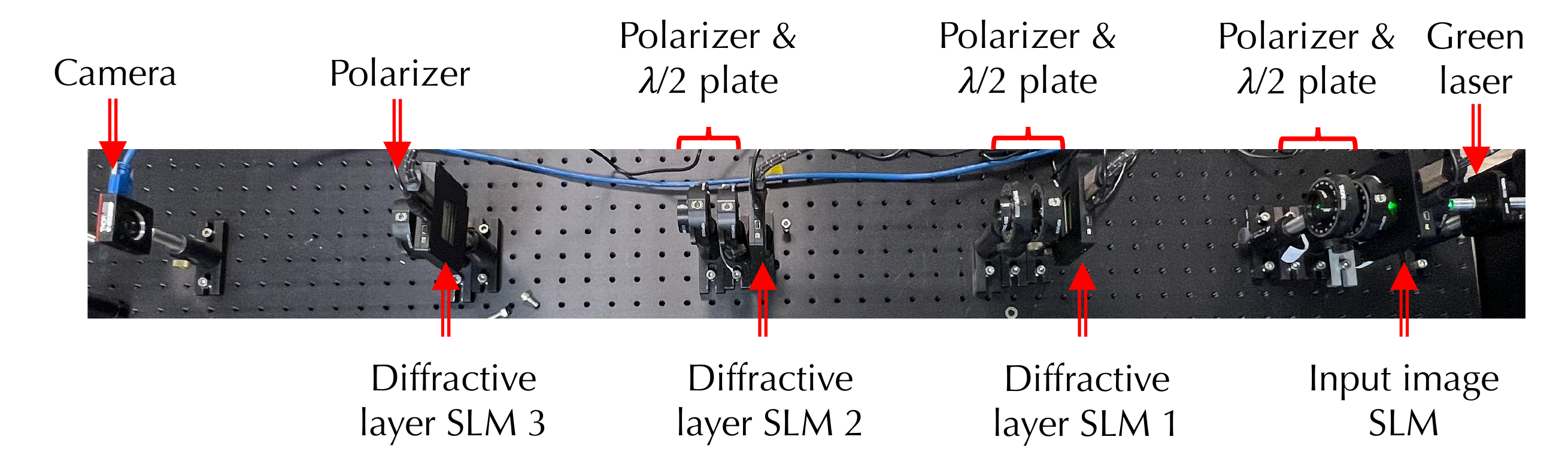}
	\end{center}
	\caption{\small \textbf{Photo of the all-optical reconfigurable DONNs experimental system.} The system consists of a green laser, four spatial light modulators (SLMs), multiple polarizers and half-waveplates ($\lambda$/2 plate), and a camera.}
\end{figure}

\newpage
\begin{figure}[h]
	\begin{center}
		\includegraphics[width=\textwidth]{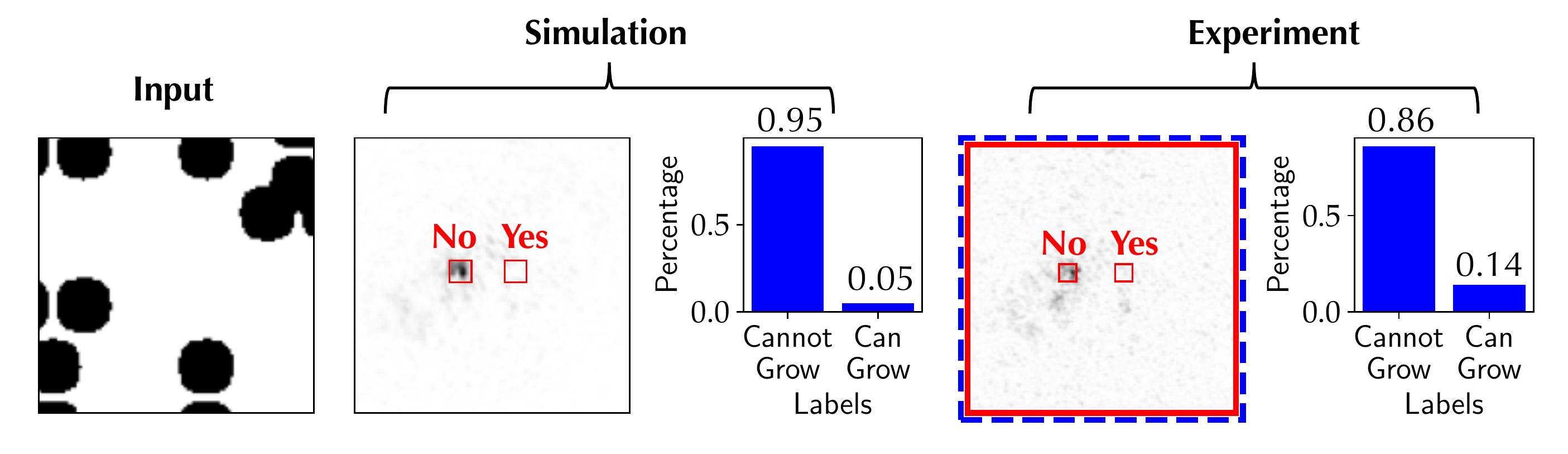}
	\end{center}
	\caption{\small \textbf{Additional data on the prediction of 2D quantum materials synthesis.} Input image, simulation camera output image and corresponding detector intensity distribution, and experimental camera output image and corresponding detector intensity distribution are shown.}
\end{figure}

\newpage
\begin{figure}[h]
	\begin{center}
		\includegraphics[width=\textwidth]{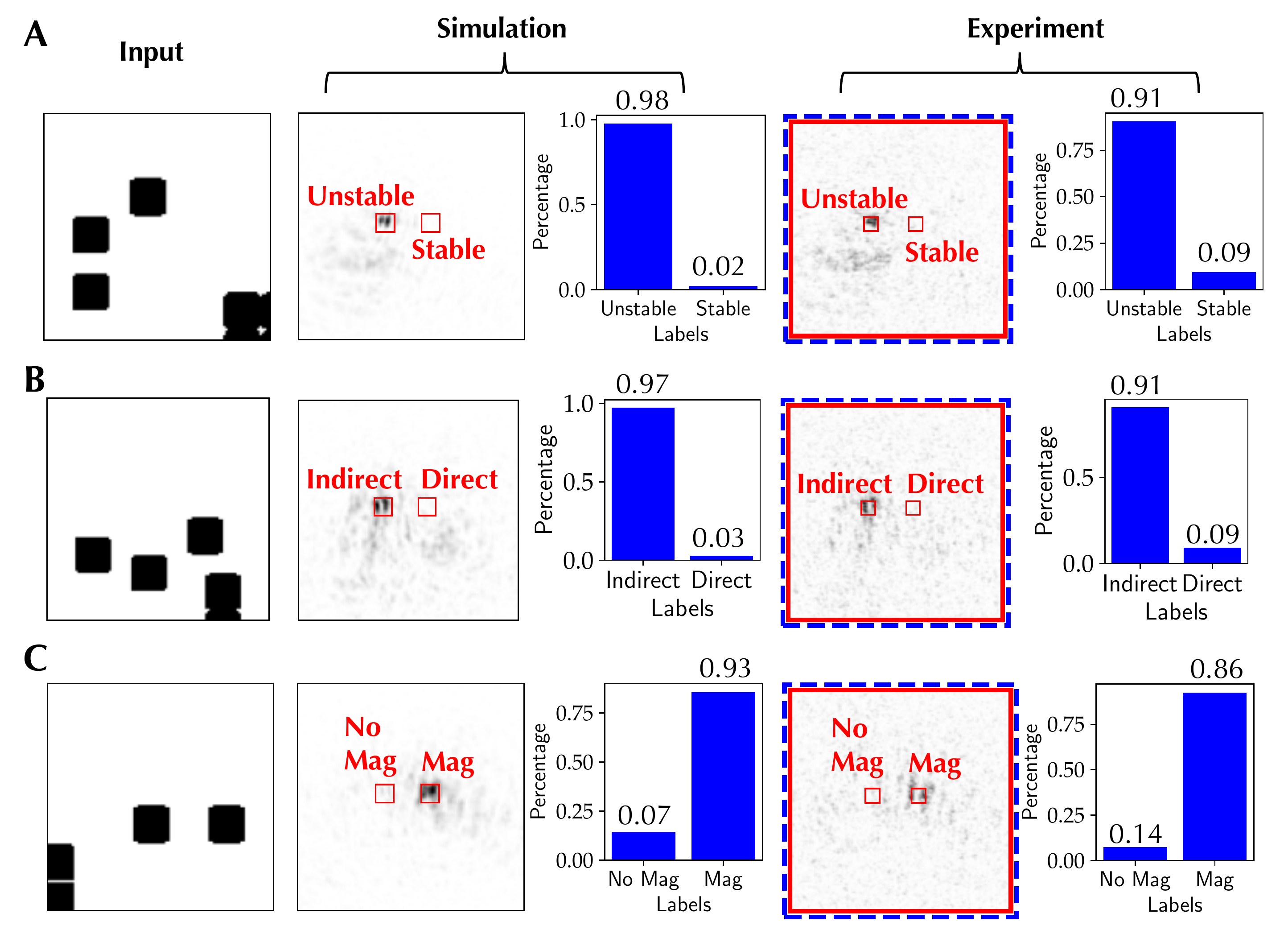}
	\end{center}
	\caption{\small \textbf{Additional data on the prediction of 2D quantum material properties.} Input images, simulation camera output images and corresponding detector intensity distributions, and experimental camera output images and corresponding detector intensity distributions are shown for predicting the \textbf{(A)}\,stability, \textbf{(B)}\,band structure, and \textbf{(C)}\,magnetism of 2D quantum materials.}
\end{figure}

\newpage
\begin{figure}[h]
	\begin{center}
		\includegraphics[width=\textwidth]{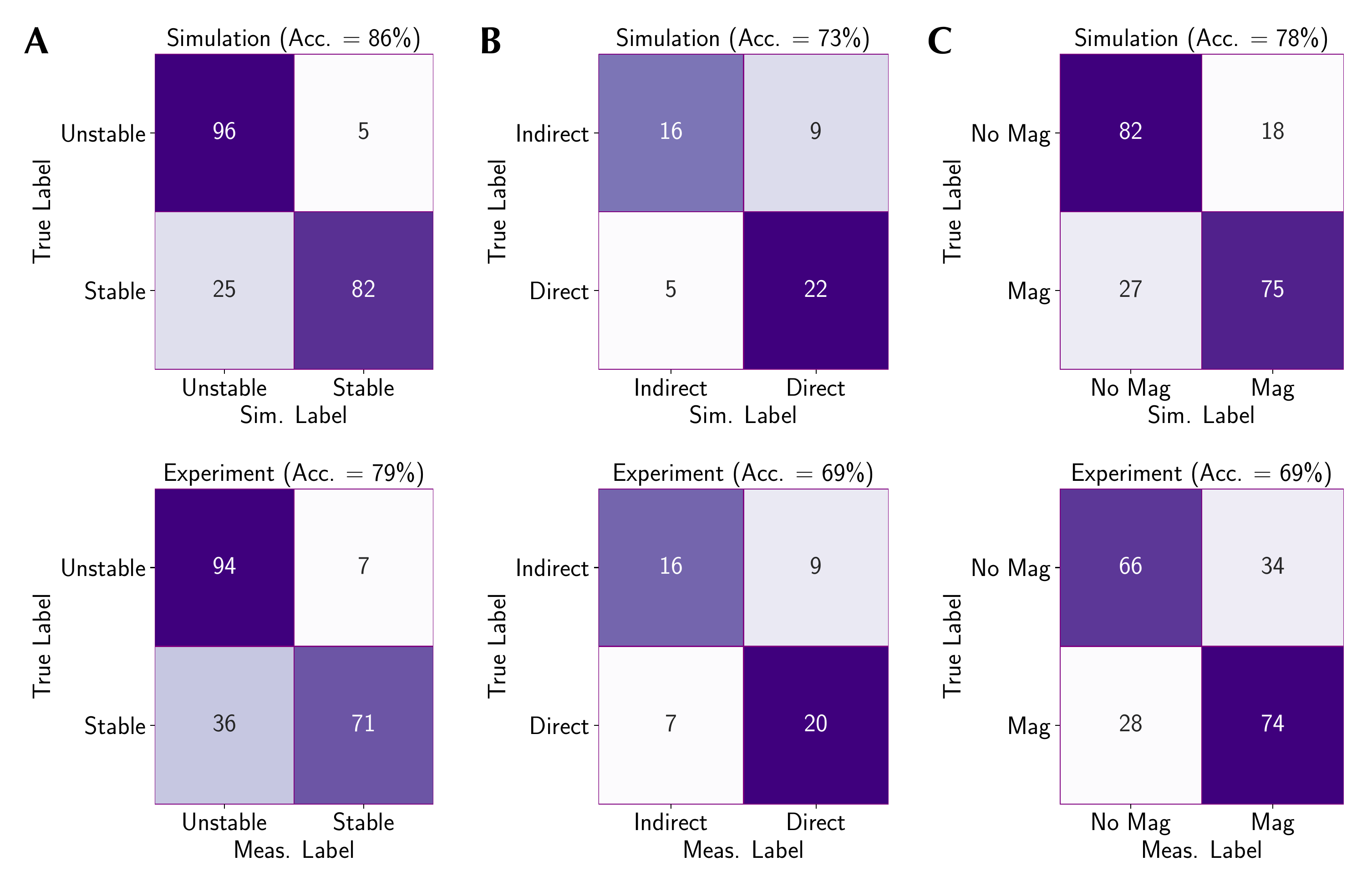}
	\end{center}
	\caption{\small \textbf{Confusion matrices for the prediction of 2D quantum material properties.} Simulation and experimental confusion matrices for predicting \textbf{(A)}\,the stability, \textbf{(B)}\,band structure, and \textbf{(C)}\,magnetism.}
\end{figure}

\newpage
\begin{figure}[h]
	\begin{center}
		\includegraphics[width=\textwidth]{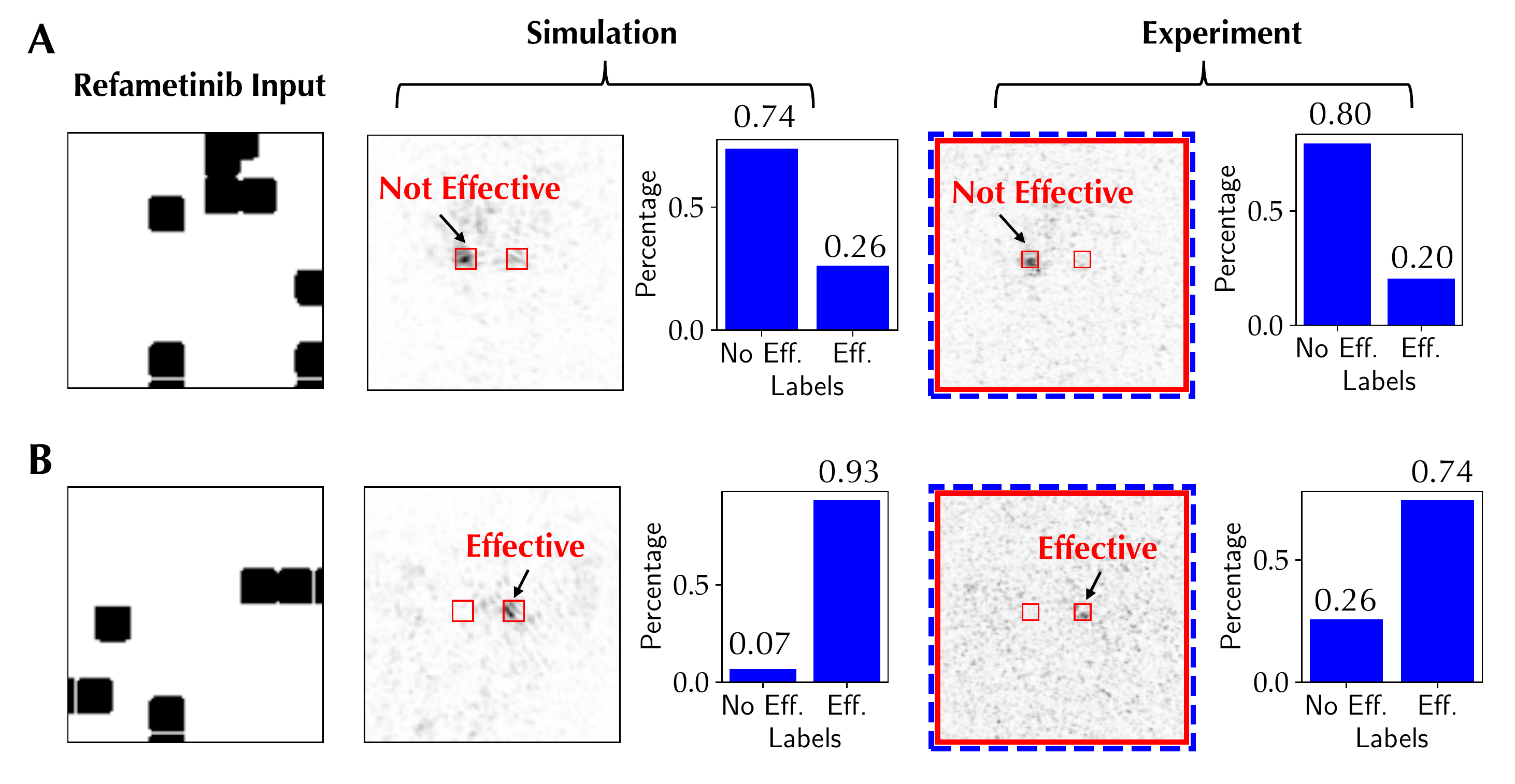}
	\end{center}
	\caption{\small \textbf{Effectiveness prediction of Refametinib.} Input images, simulation camera output images and corresponding detector intensity distributions, and experimental camera output images and corresponding detector intensity distributions for \textbf{(A)}\,an ineffective case and \textbf{(B)}\,an effective case.}
\end{figure}

\newpage
\begin{figure}[h]
	\begin{center}
		\includegraphics[width=\textwidth]{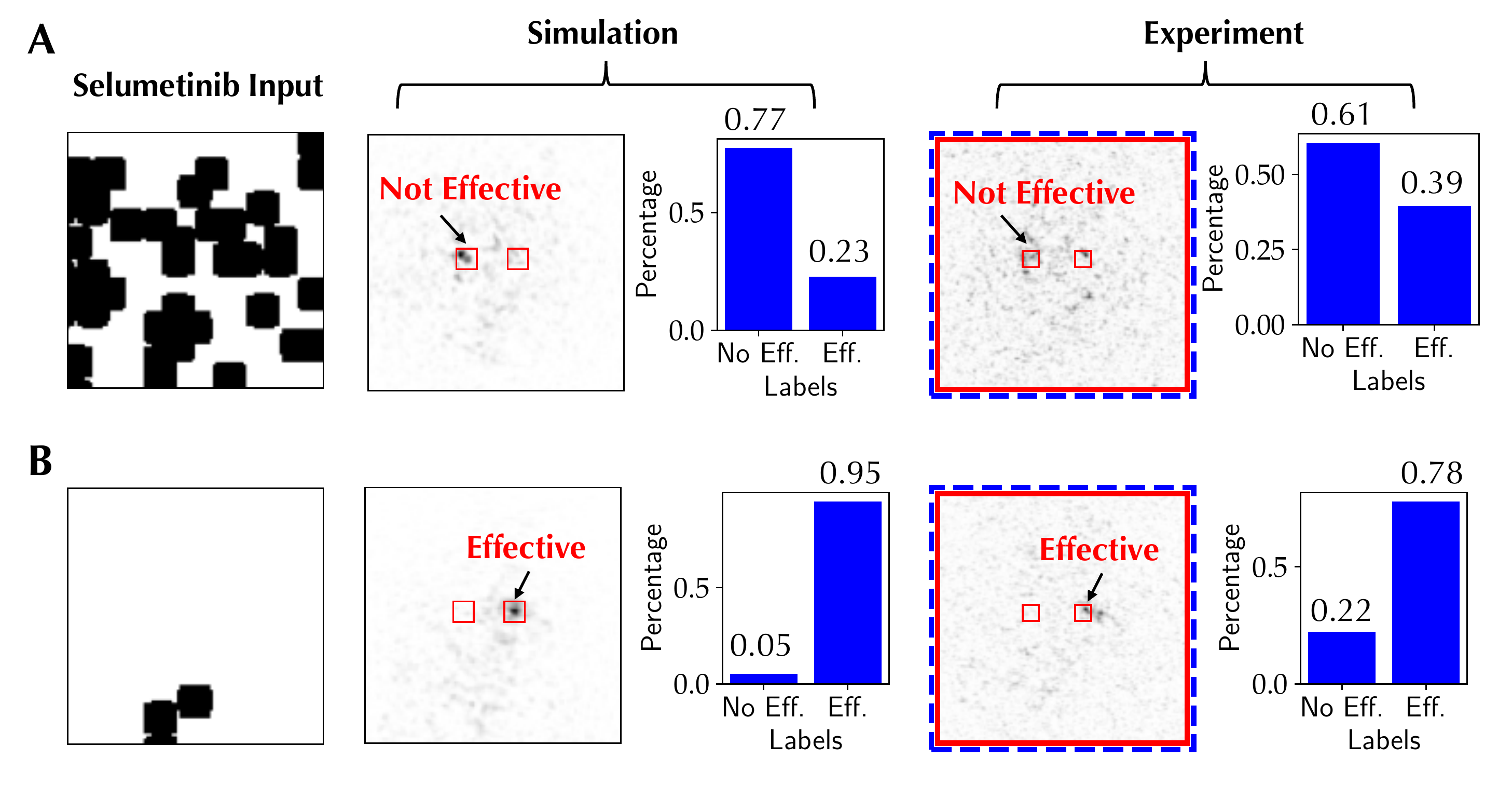}
	\end{center}
	\caption{\small \textbf{Effectiveness prediction of Selumetinib.} Input images, simulation camera output images and corresponding detector intensity distributions, and experimental camera output images and corresponding detector intensity distributions for \textbf{(A)}\,an ineffective case and \textbf{(B)}\,an effective case.}
\end{figure}

\newpage
\begin{figure}[h]
	\begin{center}
		\includegraphics[width=\textwidth]{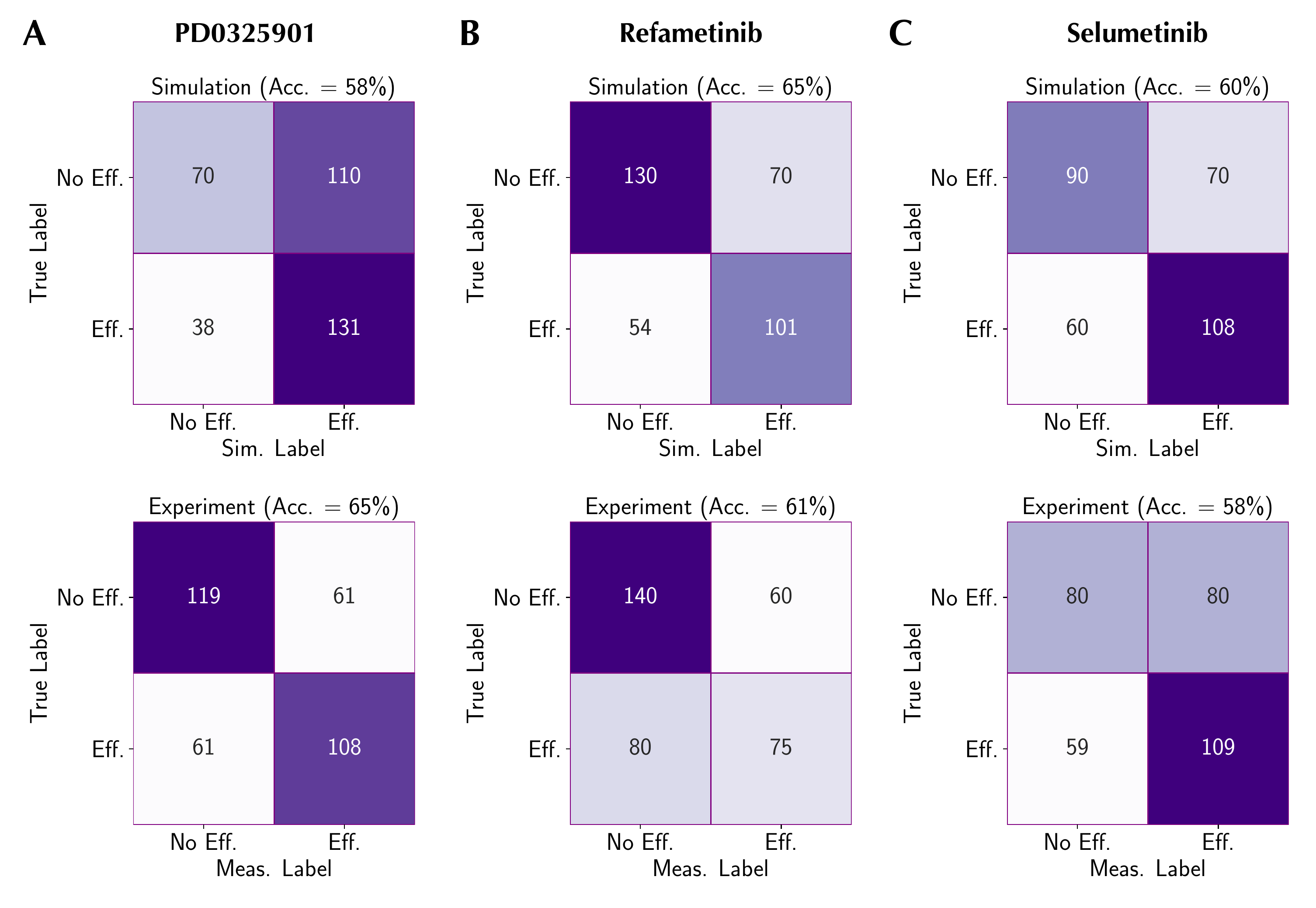}
	\end{center}
	\caption{\small \textbf{Confusion matrices for the prediction of cancer inhibitor effectiveness.} Simulation and experimental confusion matrices for the effectiveness prediction of \textbf{(A)}\,PD0325901, \textbf{(B)}\,Refametinib, and \textbf{(C)}\,Selumetinib.}
\end{figure}

\newpage
\begin{figure}[h]
	\begin{center}
		\includegraphics[width=\textwidth]{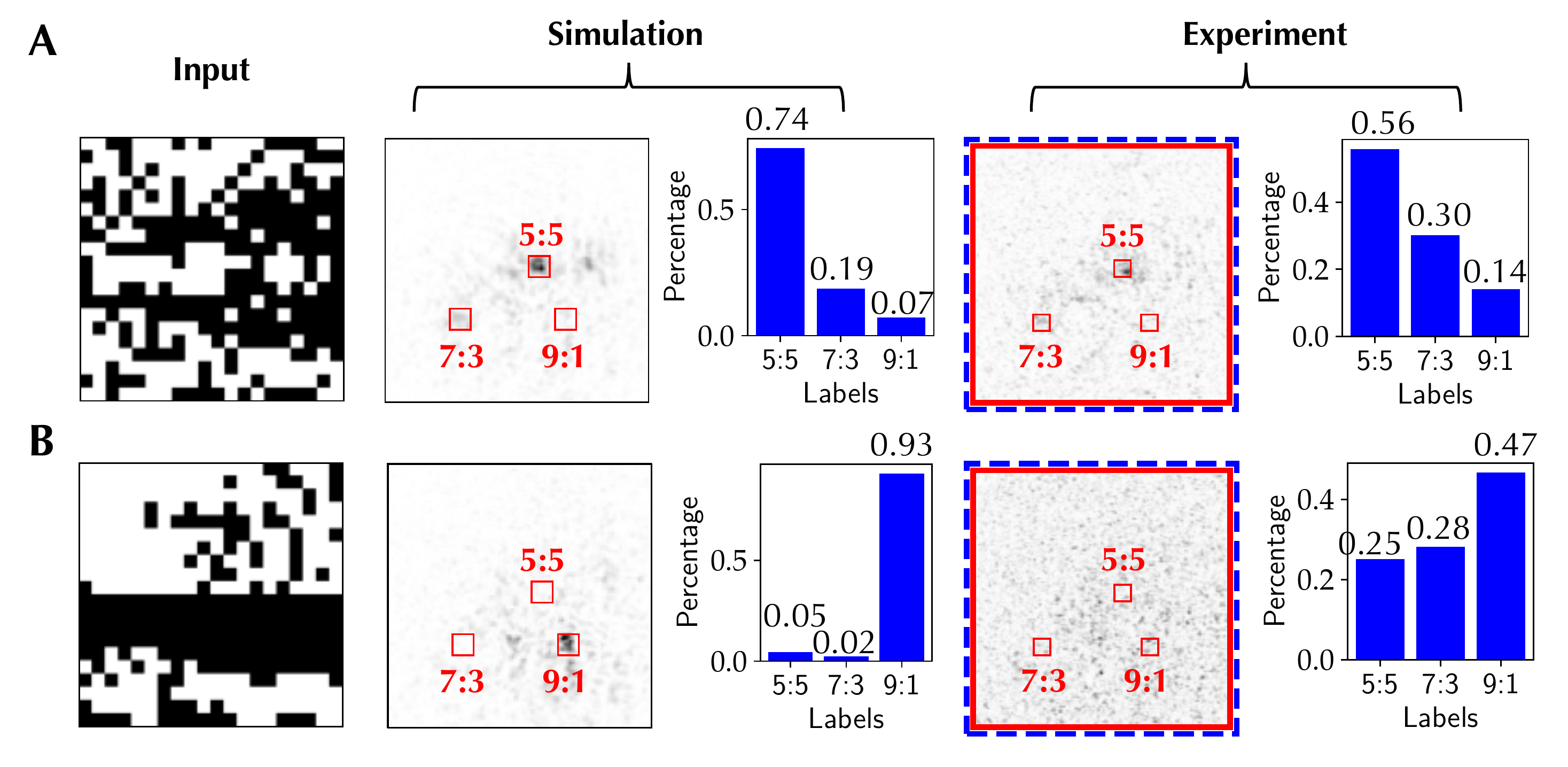}
	\end{center}
	\caption{\small \textbf{Additional data on the prediction of the nanopatterned integrated photonic power splitter.} Input images, simulation camera output images and corresponding detector intensity distributions, and experimental camera output images and corresponding detector intensity distributions for \textbf{(A)}\,a splitter with $5:5$ splitting ratio and \textbf{(B)}\,a splitter with $9:1$ splitting ratio.}
\end{figure}